# Branch Identification in Passive Optical Networks using Machine Learning

Khouloud Abdelli[1,3], Carsten Tropschug[2], Helmut Grießer[1], Sander Jansen[1], and Stephan Pachnicke[3]

[1] ADVA Optical Networking SE, Fraunhoferstr. 9a, 82152 Munich/Martinsried, Germany
[2] ADVA Optical Networking SE, Märzenquelle 1-3, 98617 Meiningen, Germany
[3] Christian-Albrechts-Universität zu Kiel, Kaiserstr. 2, 24143 Kiel, Germany
E-mail: KAbdelli@adva.com

**Abstract:** A machine learning approach for improving monitoring in passive optical networks with almost equidistant branches is proposed and experimentally validated. It achieves a high diagnostic accuracy of 98.7% and an event localization error of 0.5m. © 2022 The Author(s).

## 1. Introduction

Optical passive networks (PONs) have gained popularity as a promising broadband fiber access network solution due to their high bandwidth, low cost, flexibility, and scalability. PON systems are primarily deployed in fiber-to-the-home (FTTH) networks to deliver a wide range of communication and multimedia services [1]. Due to the omission of active electronic components, OPEX is reduced, and this also makes them less failure-prone in the outside plant. Implementing and deploying effective monitoring schemes in these systems can result in significant additional OPEX savings. Optical time domain reflectometry (OTDR), a technique based on Rayleigh backscattering, has primarily been used to monitor individual optical fiber spans. However, applying OTDR to monitor PON systems can be challenging because the backscattered signals from each branch are added together, making it difficult to distinguish between the backward signals of the individual branches [2]. In the case of (almost) equidistant branch terminations, event analysis becomes most difficult as the reflected signals from the branches with the same length overlap and add up. Furthermore, the high loss of the optical splitters at the remote node causes a significant drop in the backscattered signal (e.g., an 1:32 splitter leads to 15 dB loss in measured power [3]). To address the aforementioned challenges, one proposed solution is to use a tunable OTDR in conjunction with wavelength multiplexers to provide a dedicated monitoring wavelength for each branch. However, such a solution is quite expensive due to the high cost of a tunable OTDR network device. Another simple and effective solution is to use optical reference reflectors at the end of each branch to improve distinguishability by detecting the presence and height variation of reference reflector peaks. However, if the lengths of the branches are almost identical, the conventional OTDR event detection methods used to analyze the data derived from the reflectors fail to perform well, limiting their applicability for real-world installed networks with similar branch lengths. Recently, a machine learning (ML) based approach for fault diagnosis in PONs [4] has been proposed, but the focus there was to determine the fault type in a PON system rather than to identify the faulty branch.

To overcome the aforementioned shortcomings, we propose in this paper a ML-based approach for event monitoring in PON systems with similar branch lengths, whereby the branches can be identified even when the reflected pulses overlap, by leveraging monitoring data obtained from the reflectors installed at each branch's end. The efficiency of the proposed method is validated using experimental OTDR data derived from a PON system. The robustness and generalization ability of the ML approach are evaluated using previously unseen experimental data from a different PON system.

## 2. Proposed Approach

In our implementation, the OTDR traces recorded from a PON system are firstly split into fixed-length sequences, and the ML model is then applied to each obtained sequence to simultaneously output the event type ($C_0$: no reflection, $C_1$: one reflection, and $C_2$: two reflections) and its location(s). The length of the sequence to be fed into the ML model is chosen in such a way that the likelihood of more than two reflections within that sequence for the pulse width used in the OTDR measurements is extremely low. It should be noted that the two reflection patterns considered for the training and inference phases include overlapped peaks where one of the reflections is partially or completely obscured, making them difficult to distinguish.

### 2.1 Experimental Data

The experimental setup A shown in Fig. 1 is used to record OTDR traces incorporating various reflection patterns under different conditions. A real passive optical network is reproduced by using a cascade of optical splitters with a splitting ratio of 1:32. A reflector or an open physical contact (PC) connector is placed at the end of each branch to induce a reflection. The height of each generated reflection is adjusted using a static optical attenuator with various attenuation settings ranging from 1 to 16 dB, in order to produce diverse patterns, including cases representing faulty branch conditions. The length difference between the neighboring branches (the second and third branches) is varied from 1 to 3 m to generate two reflection samples with overlapped reflected pulses. A variable optical attenuator (VOA) located after 1000 m fiber is used to simulate the effect of a longer feeder fiber or a higher splitting ratio. VOA attenuation can range between 0 and 12 dB. The OTDR configuration parameters, namely the pulse width, and the wavelength are set to 10 ns, and 1650 nm, respectively. The laser power is varied from 0 to 12 dBm. The averaging time for collecting and averaging the OTDR traces ranges from 2 ms to 2 s. In total, ~8,159 OTDR traces are recorded, and normalized. For each generated OTDR trace and investigated event type, four sequences of length 50, including the pattern of that event, are extracted at random. Please note that the location of the investigated event type (i.e., one or two reflections) within the derived sequences is varied by changing the start point of the OTDR trace split, in order to improve the predictive power of the ML model in practice, where the ML model is applied to arbitrarily split OTDR sequences that may include the event type of interest at different locations. For each derived sequence, the event type ($C_i$ $\forall i \in \{0 \ldots 2\}$), and the event position index/indices within the sequence are assigned. A dataset composed of 97,902 samples (~32,634 examples for each investigated event type) is built, and divided into a training (60%), a validation (20%) and a test dataset (20%).

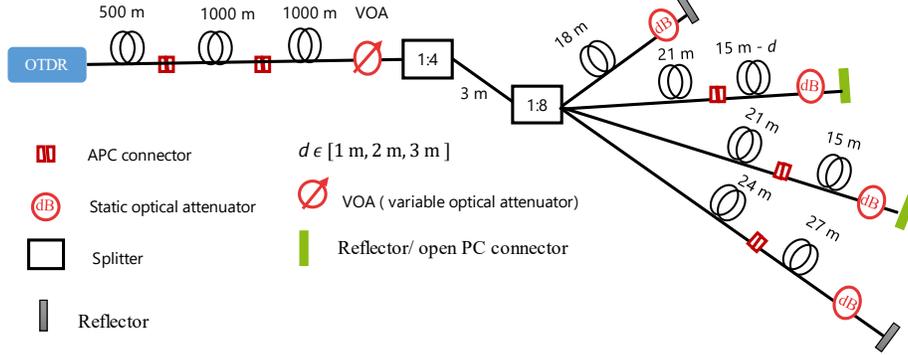

Figure 1: Experimental setup A for generating different reflection patterns in passive optical network systems.

## 2.2 Model Architecture

The structure of the proposed ML approach is shown in Fig. 2. The ML model takes as input an OTDR sequence of length 50 and outputs both the event type and its location. The input is firstly fed into a gated recurrent unit (GRU) [5] layer with 64 cells to capture the relevant sequential features $[h_1, h_2, \ldots h_{50}]$. The attention layer assigns then to each extracted feature $h_i$ a weight (i.e., attention score) $\alpha_i$ [6]. Afterwards, the different computed weights $\alpha_i$ are aggregated to produce a weighted feature vector (i.e., attention context vector) $c$, computed as $\sum_i \alpha_i h_i$, which captures the relevant information to improve the performance of the model. Following that, $c$ is transferred to two task-specific layers, each with 16 neurons dedicated to solving the tasks of reflection type classification ($T_1$) and event localization ($T_2$), respectively, by leveraging the knowledge extracted by the attention-based GRU shared layers. The ML model's total loss is calculated as the weighted sum of the two individual task losses, whereby the cross-entropy loss and the regression loss (mean square error) are the losses of $T_1$ and $T_2$, respectively.

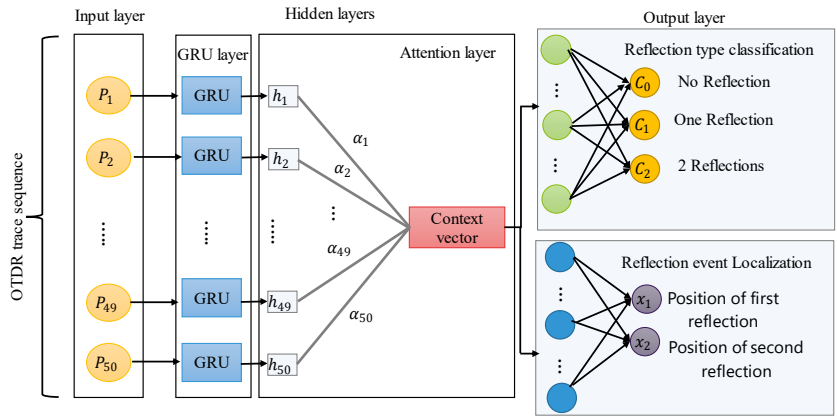

Figure 2: Proposed ML model architecture.

## 3. Results and Discussions

The performance of the ML approach is firstly evaluated using an unseen test dataset. The confusion matrix depicted in Fig. 3 (a) shows that the ML approach classifies the different classes (i.e., event types) by achieving a good average diagnostic accuracy of 98.7%. In rare cases, the classes $C_1$ and $C_2$ may be misclassified, particularly for the instances of class $C_2$ where one of the reflections has vanished completely. Figure 3 (b) demonstrates that the ML approach achieves very small prediction errors (representing the difference between the predicted and actual position of the investigated event type), with a mean of 0.05 m and a standard deviation of 0.6, proving that the ML model accurately localizes the events. Figure 3 (c) shows the effect of increasing VOA attenuation (which has the same effect as increasing the length of the feeder fiber or the splitting ratio) on the diagnostic accuracy. The diagnostic accuracy remains high (greater than 96.5%) even for longer feeder fiber (up to 62.5 km) or higher splitting ratio.

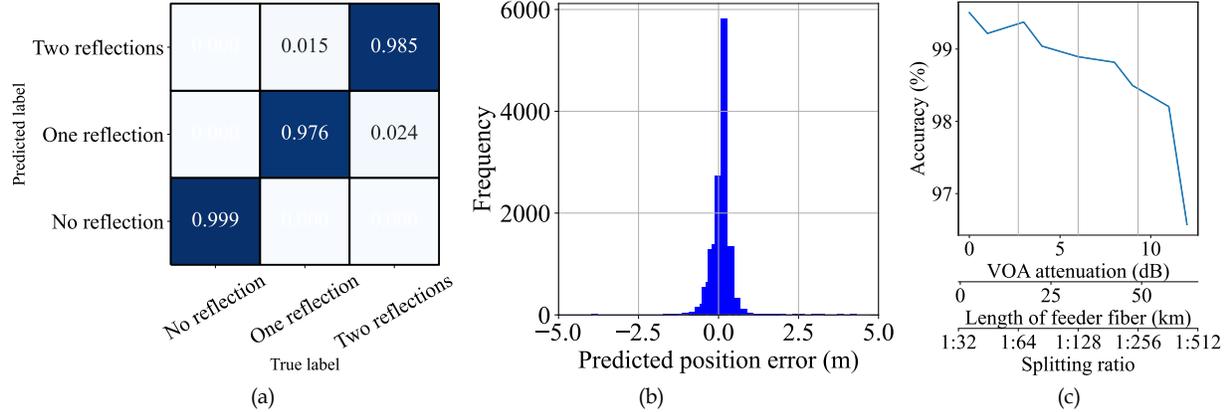

Figure 3: (a) Confusion matrix, (b) Histogram of the event position prediction errors, and (c) Effect of VOA attenuation on accuracy.

The proposed ML approach is compared to the current state-of-the-art ML techniques, namely multilayer perceptron (MLP), convolutional neural network (CNN), long-short term memory (LSTM), and GRU, in terms of diagnostic accuracy and root mean square error (RMSE). The proposed method outperforms the other ML algorithms, as shown in Tab.1, by achieving the highest accuracy and lowest RMSE metric thanks to the combination of the attention mechanisms and GRU, which helps to better capture relevant patterns in the data.

The performance of the ML model is then evaluated given unseen test data derived from a different experimental setup B depicted in Fig. 4, to test the robustness of the proposed approach. The ML approach is compared to a conventional method, a threshold-based system, in which the threshold chosen for detecting a reflection is optimized given the test dataset to improve the detection capability and reduce the false alarm rate. The results shown in Fig. 5 demonstrate that, first, the proposed model has a good generalization capability (yielding high accuracy when tested with previously unseen data), and that, second, our approach outperforms the conventional method significantly. The conventional method performs worse because it fails to identify overlapped reflections where the second reflection disappears entirely or partially due to the high attenuation settings.

Table 1. Comparison of different ML methods

| Method | Accuracy (%) | RMSE (m) |
|---|---|---|
| MLP | 77.3 | 2.3 |
| CNN | 86.9 | 2.04 |
| LSTM | 96.2 | 1.1 |
| GRU | 97.9 | 0.74 |
| Proposed method | **98.6** | **0.5** |

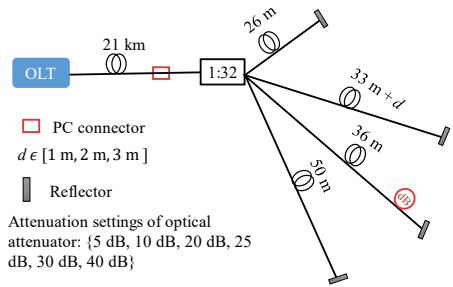
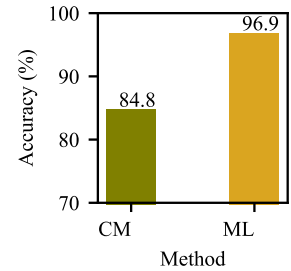

Fig. 4: Experimental setup B for generating test data.  Fig. 5: Comparison of different approaches.

## 4. Conclusion

We proposed and experimentally validated an ML-based approach for enhancing fiber branch monitoring in PON systems with almost equidistant branches. The experimental results have proven that the proposed model outperforms other cutting-edge ML techniques as well as the conventionally employed method. In the future, we plan to generalize the ML model to identify more than two reflections (i.e., branches) within a sequence.

This work has been performed in the framework of the CELTIC-NEXT project AI-NET PROTECT (Project ID C2019/3-4), and it is partly funded by the German Federal Ministry of Education and Research (FKZ16KIS1279K).